\documentclass[onecolumn]{article}          
\usepackage[pdftex]{graphicx}
\usepackage{amsmath}
\usepackage{amssymb}
\usepackage{epsfig}
\usepackage{epstopdf}
\usepackage{color}
\usepackage{subfigure}
\usepackage{authblk}
\usepackage{algorithm}
\usepackage{algorithmic}
\usepackage{array}
\usepackage{multirow} 
\usepackage{caption}
\usepackage{mathtools}
\usepackage{siunitx}

\DeclarePairedDelimiter{\ceil}{\lceil}{\rceil}

\DeclareMathOperator*{\argmin}{arg\,min}

\newcommand{\real}{\mathbb{R}}

\def\thesection{\arabic{section}}
\renewcommand{\thetable}{\arabic{table}}
\renewcommand{\thefigure}{\arabic{figure}}

\begin{document}
\sloppy
\title{Geometry-based Symbolic Approximation for Fast Sequence Matching on Manifolds}

\author{Rushil Anirudh}
\author{Pavan Turaga}
\affil{ School of Electrical, Computer and Energy Engineering\\School of Arts, Media and Engineering\\Arizona State University, Tempe AZ.}

\date{}

\maketitle

\begin{abstract}
 In this paper, we consider the problem of fast and efficient indexing techniques for sequences evolving in non-Euclidean spaces. This problem has several applications in the areas of human activity analysis, where there is a need to perform fast search, and recognition in very high dimensional spaces. The problem is made more challenging when representations such as landmarks, contours, and human skeletons etc. are naturally studied in a non-Euclidean setting where even simple operations are much more computationally intensive than their Euclidean counterparts. We propose a geometry and data adaptive symbolic framework that is shown to enable the deployment of fast and accurate algorithms for activity recognition, dynamic texture recognition, motif discovery. Toward this end, we present generalizations of key concepts of piece-wise aggregation and symbolic approximation for the case of non-Euclidean manifolds. We show that one can replace expensive geodesic computations with much faster symbolic computations with little loss of accuracy in activity recognition and discovery applications. The framework is general enough to work across both Euclidean and non-Euclidean spaces, depending on appropriate feature representations without compromising on  the ultra-low bandwidth, high speed and high accuracy. The proposed methods are ideally suited for real-time systems and low complexity scenarios.
\end{abstract}

\section{Introduction}
In this paper we consider the problem of fast comparison of sequences of structured visual representations, which have non-Euclidean geometric properties. Examples of such structured representations include 
shapes \cite{Kendall1984,Srivastava2011Elastic}, optical flow \cite{chaudhrycvpr2009}, covariance matrices \cite{Tuzel2006} where underlying distance metrics are highly involved and even simple statistical operations are usually iterative. 

Utilizing Riemannian geometric concepts have resulted in many advances in understanding complex representations. For example, features such as contours \cite{JoshiKSJ07}, skeletons \cite{VemulapalliCVPR2014}, the space of $d \times d$ covariance matrices or tensors which appear both in medical imaging \cite{Pennec2006Tensor} as well as texture analysis \cite{Tuzel2006} etc., have proven effective in image analysis. In video analysis, techniques have included temporal information using Riemannian properties such as, 
video modeling by linear dynamic systems \cite{TuragaPAMI2011}, and tensor decomposition \cite{LuiCVPR2010} etc. Long-term complex activities are often modeled as time-varying linear dynamical systems \cite{TuragaCVPR2009}, which can be interpreted as a sequence of points on a Grassmann manifold, motivating application for the problem of indexing of manifold sequences.

For these manifolds, standard notions of distance, statistics, quantization etc. need modification to account for the non-linearity of the underlying space. As a result, basic computations such as geodesic distance, finding the sample mean etc. are highly involved in terms of computational complexity, and often result in iterative procedures further increasing the computational load making them impractical. To address this issue, in this paper we propose a geometry-based symbolic approximation framework, as a result of which low-bandwidth transmission and accurate real-time analysis for recognition or searching through sequential data become fairly straightforward. 

We propose a framework that generalizes a popular indexing technique used to mine and search for vector space time series data known as Symbolic Aggregate Approximation (SAX) \cite{LinKLC03} to Riemannian manifolds. To the best of our knowledge, we are the first to propose such an indexing scheme for manifold sequences. The main idea is to replace manifold sequences with abstract {\it symbols} or {\it prototypes}, that can be learned offline. Symbolic approximation is a combination of discretization and quantization on manifold spaces, which allows us to approximate distance metrics between sequences in a quick and efficient manner. Another advantage is extremely fast searching that is possible because the search is limited to the symbolic space. Further, to enable efficient searching techniques, we develop prototypes or symbols which divide the space into equiprobable regions by proposing the first manifold generalization of a conscience based competitive learning algorithm \cite{Desieno1988}. Using these prototypes, we demonstrate that signals or sequences on manifolds can be approximated effectively such that the resulting metric remains close to the metric on the original feature space, thereby guaranteeing accurate recognition and search. While this framework is applicable to general high-dimensional feature sequences, we demonstrate its utility on a few common video-analysis problems such as activity analysis and dynamic texture modeling. Generally speaking, the ideal symbolic representation is expected to have two key properties: (1) be able to model the data accurately with a low approximation error, and (2) should enable the efficient use of existing data structures and algorithms, developed for string searching. 

\noindent We summarize our contributions next. 

\vspace{-5pt}
\paragraph{Contributions: }
\begin{enumerate}
\item We present a geometry based data-adaptive strategy for indexing time series evolving on non-Euclidean spaces. We demonstrate the effectiveness on three manifolds namely the hypersphere, the Grassmann manifold and the product space of $SE(3)\times \dots \times SE(3)$.
\item We propose the first generalization of competitive learning algorithms to Riemannian manifolds for this task, such that they are able learn prototypes which enable efficient searching.
\item The resulting framework allows the comparison between two manifold sequences at speeds nearly $100 \times$ faster than geodesic based comparisons. 
\item Applications in activity recognition and discovery show that the speed up can be achieved with minimal loss of accuracy as compared to the original features. 
\end{enumerate}

\paragraph{Organization: }
In Sec \ref{sec:related}, we discuss works related to indexing on non-Euclidean and Euclidean spaces. Next, Sec \ref{sec:manifoldsPrelim} introduces the manifolds used in this paper namely - Grassmann, Hypersphere, and the product space of $SE(3)$, including their geometric properties. Sec \ref{sec:manifold-sax} introduces the extension of SAX to Riemannian manifolds, which includes the generalization of conscience based competitive learning in algorithm \ref{Alg:Conscience}. Sec \ref{sec:speedup} presents the application of string-based algorithms to speedup search and discovery of manifold sequences, applied to human activities. Finally, Sec \ref{sec:expt} discusses the experiments on different manifold valued features on publicly available activity datasets. We conclude the paper and discuss extensions and generalizations in Sec \ref{sec:discuss}.

\section{Related Work}
\label{sec:related}
\paragraph{Indexing static points on non-Euclidean spaces}
Not surprisingly, many standard approaches for sequence modeling and indexing which are designed for vector-spaces need significant generalization to enable application to non-Euclidean spaces. Indexing of static data on manifolds has been addressed recently with hashing based approaches \cite{Chaudhry2010}. For data points lying on the space of Symmetric Positive Definite (SPD) matrices, \cite{HarandiSH14} present a dimensionality reduction technique that is geometry aware. Our interest lies in indexing sequences directly instead of individual points. Signal approximation for manifolds using wavelets \cite{RahmanSIAM05} is a related technique. However, it is non-adaptive to the data and requires observing the entire signal before it can be approximated, while the proposed framework allows for easy real time implementation once the symbols are learned. Recent work also dealt with modeling human activity as a manifold valued random process \cite{YiKN12} where the proposed techniques are theoretically and computationally involved due to the requirement of second-order properties such as parallel transports. Another related line of work in recent years has been advances in Riemannian metrics for sequences on manifolds \cite{Srivastava2011Elastic}. These approaches consider a sequence as an equivalent vector-field on the manifold. A distance function is imposed on such vector-fields in a square-root elastic framework. This is applied to the special case of curves in $2D$, $nD$, and non-Euclidean spaces \cite{Srivastava2011Elastic,JoshiKSJ07,Jingyong2014}. While such a distance function could be utilized for the purposes of indexing and approximation of sequences, it is offset by the computational load required in computing the distance function for long sequences.

\paragraph{Computationally efficient representations of images and video}
In the past decade, there has been significant progress in efficient retrieval and indexing techniques \cite{ChumPM09} for very large image datasets. There have also been extensions to video retrieval \cite{RevaudDSJ13} from very large databases. These techniques have made it possible to search accurately through large image and video data bases, but most methods are for high dimensional Euclidean points or time-series, and their generalization for manifold valued data is unclear. 

\paragraph{Euclidean time-series indexing}
A successful approach to tackle the problem of fast indexing of {\em scalar} sequences has been to discretize and quantize the sequence in a way such that the obtained symbolic form contains most of the information of the original sequence, yet enabling much faster computations.
 This class of approaches are broadly termed as {\bf S}ymbolic {\bf A}ggregate Appro{\bf x}imation (SAX) \cite{LinKLC03}. Several problems of indexing and motif discovery from time series have been addressed using this framework  \cite{LinKLC03,MueenKZCW09},
however the extension from $1D$ to multidimensional and non Euclidean spaces is not trivial. Multidimensional extensions to SAX have also been proposed such as \cite{VahdatpourAS09}, but these are trivial extensions which perform SAX on every dimension individually without considering the  geometry of the ambient space.

Further, for manifolds such as the Grassmannian or the function-space of closed curves, there is no natural embedding into a vector space, thus motivating the need for a geometry-based intrinsic approach \cite{Spivak1999,Srivastava2011Elastic}. We show that this class of approaches can be generalized to take into account the geometry of the feature space resulting in several appealing characteristics, as they enable us to replace highly non-linear distance function computations with much faster and simpler symbolic distance computations. 

\paragraph{Efficient string searching}
The biggest advantage of using the proposed indexing method is the the representation of complex feature types using abstract symbols, that are learned offline. This enables the use of string searching algorithms, allowing one to search through very high dimensional, non-linear spaces with a $O(m+n)$ complexity or better, where $m$ and $n$ are the length of a query, and the size of a activity database respectively. A known result in data mining is that the computational complexity can be further reduced to $O(m + n (log_{|\Sigma|}m)/m)$, for an alphabet of size $\Sigma$, when the symbols are independent and equiprobable \cite{Allauzen2000}. Other lower bounds have been proposed when symbols are equiprobable \cite{Yao1979}, and it is known the height of suffix trees is optimized with equiprobable symbols \cite{devroye92}. The vector space SAX \cite{LinKLC03} proposed to generate symbols by partitioning the Gaussian distribution into bins of equal probability. However, it is not trivial to partition the data space into equiprobable regions on manifolds hence we use a conscience based competitive learning algorithm to learn the codebook.

\section{Mathematical Preliminaries}
\label{sec:manifoldsPrelim}
\label{sec:manifoldsPrelim}
In this section we will outline the geometric properties of the manifolds considered in this work, namely the Grassmann, hyper-sphere and the space of $SE(3)\times \dots SE(3)$. For an overview on Riemannian geometry and topology, we refer the readers to useful resources on the topic \cite{Absil2004,Boo03}. Next we describe the different features and their respective geometric spaces.

\paragraph{\bf Landmarks on the Silhouette:} We represent a shape as a $m \times 2$ matrix $L = [(x_1,y_1)$;$(x_2,y_2);\ldots ; (x_m,y_m)]$, of the set of $m$ landmarks of the zero-centered shape. The \emph{affine shape space} \cite{Goodall1999} is useful to remove the effects of small variations in camera location or small changes in the pose of the subject. Affine transforms of the base shape $L_{base}$ can be expressed as $L_{affine}(A) = L_{base} * A^{T}$, and this multiplication by a full-rank matrix on the right preserves the column-space of the matrix $L_{base}$. Thus, the 2D subspace of  $\mathbb{R}^m$ spanned by the columns of the matrix $L_{base}$ is an \emph{affine-invariant} representation of the shape. i.e. $span(L_{base})$ is invariant to affine transforms of the shape. Subspaces such as these can be identified as points on a Grassmann manifold \cite{TuragaPAMI2011}.

A given $d$-dimensional subspace of $\mathbb{R}^m$, $\mathcal{Y}$ can be associated with a idempotent rank-$d$ projection matrix $P = YY^T$, where $Y$ is a $m \times d$ orthonormal matrix such as $span(Y) = \mathcal{Y}$. The space of $m \times m$ projectors of rank $d$, denoted by $\mathbb{P}_{m,d}$ can be embedded into the set of all $m \times m$ matrices - $\real^{m \times m}$- which is a vector space. Using the embedding $\Pi:  \mathbb{R}^{m\times m} \rightarrow \mathbb{P}_{m,d}$  we can define a distance function on the manifold using the metric inherited from $\real^{m \times m}$.

\begin{equation}
\label{eq:grass_dist}
d^2(P_1,P_2) = tr(P_1-P_2)^T(P_1-P_2)
\end{equation}
The distance metric defined in \eqref{eq:grass_dist} is closely related to the Procrustes measure on the Grassmann manifold which has previously been used in \cite{CetingulV09}.

The projection $\Pi:  \mathbb{R}^{m\times m} \rightarrow \mathbb{P}_{m,d}$ is given by:

\begin{equation}
\label{proj_func}
\Pi(M)=UU^T
\end{equation}
where  $M = USV^T$ is the $d$-rank SVD of M. \\ \\

Given a set of sample points on the Grassmann manifold represented uniquely by projectors \{$P_1,P_2,...P_N$\}, we can compute the extrinsic mean \cite{TuragaChapter2010} by first computing the mean of the $P_i$'s and then projecting it to the manifold as follows :
\begin{equation}
\mu_{ext}=\Pi(P_{avg}), \mbox{where }P_{avg}=\frac{1}{N}\sum^N_{i=1}P_i
\end{equation}

\paragraph{\bf Histograms of Oriented Optical Flow (HOOF):} As described in \cite{chaudhrycvpr2009}, optical flow is a natural feature for motion sequences. Directions of Optical Flow vectors are computed for every frame, then binned according to their primary angle with the horizontal axis and weighted according to their magnitudes. Using magnitudes alone is susceptible to noise and can be very sensitive to scale. Thus all optical flow vectors, $v = [x,y]^T$ with direction $\theta = tan^{-1}(\frac{y}{x})$ in the range 
\begin{equation}
-\frac{\pi}{2}+\pi \frac{b-1}{B} \le \theta < -\frac{\pi}{2}+\pi \frac{b}{B}
\end{equation}
will contribute by $\sqrt{x^2+y^2}$ to the sum in bin $b$, $1 \le b \le B$, out of a total of $B$ bins. Finally, the histogram is normalized to sum up to 1. 
Each frame is represented by one histogram and hence a sequence of histograms are used to describe an activity. The histograms {\bf $h_t$} = [{\bf $h_{t;1}$}, \dots , {\bf $h_{t;B}$}] can be re-parameterized to the {\it square root representation} for histograms, ${\sqrt{{\bf h_t}} = [\sqrt{h_{t;1}},\dots,\sqrt{ h_{t;B}}]}$ such that ${\sum^B_{i = 1} (\sqrt{h_{t;i}})^2=1}$. The Riemannian metric between two points $R_1$ and $R_2$ on the hypersphere is  $d(R_1, R_2) = cos^{-1}(R_1^TR_2)$. This projects every histogram onto the unit B-dimensional hypersphere or $\mathbb{S}^{B-1}$. From the differential geometry of the sphere, the exponential map is defined as \cite{Srivastava2007}
 \begin{equation}
\mathrm{ exp}_{\psi_i}(\upsilon) = \mathrm{cos}(||{\upsilon}||_{\psi_i})\psi_i + \mathrm{sin}(||\upsilon||_{\psi_i})\frac{\upsilon}{||\upsilon||_{\psi_i}}
 \end{equation}
 Where $\upsilon \in T_{\psi_i}( \Psi)$ is a tangent vector at $\psi_i\mbox{ and }||\upsilon||_{\psi_i}$ = $\sqrt{\left<\upsilon,\upsilon\right>_{\psi_i}} = (\int_0^T \upsilon(s) \upsilon(s) ds)^{\frac{1}{2}}.$ In order to ensure that the exponential map is a bijective function, we restrict $||\upsilon||_{\psi_i} \in [0,\pi].$ The truncation of the domain of the the exponential map is made in accordance to the injectivity radius, which is the largest radius for which the exp map is a diffeomorphism. For the sphere, the injectivity radius is $\pi$. Points that lie beyond the injectivity radius have a shorter path connecting them to $\psi_i$, which determines their geodesic distance incorrectly. The logarithmic map from $\psi_i$ to  $\psi_j$ is given by
\begin{equation}
\overrightarrow{\psi_i\psi_j} = \mathrm{log}_{\psi_i}(\psi_j) = \frac{\textbf{ u}}{(\int_0^T \textbf{ u}(s) \textbf{ u}(s) ds)^{\frac{1}{2}}} \mathrm{cos}^{-1}\left< \psi_i, \psi_j\right>,
\end{equation}
with \textbf{ u} = $\psi_i - \left< \psi_i, \psi_j\right>\psi_j.$

\paragraph{\bf Lie Algebra Relative Pairs (LARP):}
Finally, we consider a skeletal representation proposed recently \cite{VemulapalliCVPR2014} which has been shown to be very effective for activity recognition on data obtained from depth sensors such as Microsoft Kinect. LARP represents every skeleton as a set of relative transformations between joints, where a transformation consists of a rotation and a translation and therefore lies on the Special Euclidean group $SE(3)$. Further every skeleton with $N$ joints, is represented as a set of such transformations between $ N-1 \choose 2$ relative pairs, therefore the final feature is represented as a point on a product space of $SE(3) \times \dots \times SE(3)$.

The special Euclidean group, denoted by $SE(3)$ is a Lie group, containing the set of all $4 \times 4$ matrices of the form 
\begin{equation}
P(R,\overrightarrow{d}) = \begin{bmatrix} R & \overrightarrow{d}\\0 & 1 \end{bmatrix},
\end{equation}
where $R$ denotes the rotation matrix, which is a point on the special orthogonal group $SO(3)$ and $\overrightarrow{d}$ denotes the translation vector, which lies in $\mathbb{R}^3$. The $4 \times 4$ identity matrix $I_4$ is an element of $SE(3)$ and is the identity element of the group. The exponential map, which is defined as $\mbox{exp}_{SE(3)}: \mathfrak{se}(3)\rightarrow SE(3)$ and the inverse exponential map, defined as $\mbox{log}_{SE(3)}: SE(3) \rightarrow \mathfrak{se}(3)$ are used to traverse between the manifold and the tangent space respectively. The exponential and inverse exponential maps for $SE(3)$ are simply the matrix exponential and matrix logarithms respectively, from the identity element $I_4$. The tangent space at $I_4$ of a $SE(3)$ is called the Lie algebra of $SE(3)$, denoted by $\mathfrak{se}(3)$. It is a 6-dimensional space formed by matrices of the form:
\begin{equation}
B = \begin{bmatrix}
U & \overrightarrow{w} \\
0 & 0
\end{bmatrix} = \begin{bmatrix}
0  & -u_3 & u_2 & w_1\\
u_3 & 0 & -u1 & w_2\\
-u_2 & u_1 & 0 & w_3\\
0 & 0 & 0 & 0  \\
\end{bmatrix},
\end{equation}
where $U$ is a $3 \times 3$ skew-symmetric matrix and $\overrightarrow{w} \in \mathbb{R}^3$. An equivalent representation of $B$ is $\mbox{vec}(B) = [u_1,u_2,u_3,w_1,w_2,w_3]$ which lies on $\mathbb{R}^6$.

These tools are trivially extended to the product space, the identity element of the product space is simply $(I_4,\dots,I_4)$ and the Lie algebra is $\mathfrak{m} = \mathfrak{se}(3)\times \dots \times \mathfrak{se}(3)$. 


\section{Symbolic approach for Manifold Sequences}
\label{sec:manifold-sax}

In this section, we describe the proposed representation for manifold sequences which allows efficient algorithms to be deployed for a variety of tasks such as motif discovery, low-complexity activity recognition. We focus on the piece-wise aggregate and Symbolic approximation (PAA, SAX) \cite{ChakrabartiKMP02,LinKLC03} formulation, and present an intrinsic method to extend it to non Euclidean spaces like manifolds. Briefly, the PAA and SAX formulation consist of the following principal ideas - A given 1D scalar time-series is first divided into windows and the sequence in each window is represented by its mean value. This process is referred to as {\bf piece-wise aggregation}. Then, a set of `break-points' is chosen which correspond to dividing the range of the time-series into equi-probable bins. These break-points comprise the symbols using which we translate the time series into its symbolic form. For each window, the mean value is assigned to  the closest symbol, this step is referred to as {\bf symbolic approximation}. This representation has been shown to enable efficient solutions to scalar time-series indexing, retrieval, and analysis problems \cite{LinKLC03}. 

For manifolds, to enable us to exploit the advantages offered by the symbolic representation of sequences, we need solutions to the following main problems - a) piece-wise aggregation: which can be achieved by appropriate definitions of the mean of a windowed sequence on a manifold, and b) symbolic approximation: which requires choosing a set of points that are able to represent the data well. Here, we discuss how to generalize these concepts to manifolds.
\vspace{-5pt}
\subsection{Piece-wise aggregation}
Denote the manifold of interest by $\mathcal{M}$, given a sequence $\gamma(t) \in \mathcal{M}$, we define its piece-wise approximation in terms of local-averages in small time-windows. To do this, we first need a notion of a mean of points on a manifold. Given a set of points on a manifold, a commonly used definition of their mean is the Riemanian center of mass or the Fr\'echet mean \cite{GroveK72}, which is defined as the point $\mu$ that minimizes the sum of squared-distance to all other points: 
\begin{equation}
\mu = \argmin_{x \in \mathcal{M}} \sum_{i = 1}^{N} d_{\mathcal{M}}(x,x_i)^{2},
\end{equation} 
where $d_{\mathcal{M}}$ is the geodesic distance on the manifold. 

Computing the mean is not usually possible in a closed form, and is unique only for points that are close together \cite{GroveK72}. An iterative procedure is popularly used in estimation of means of points on manifolds \cite{Pennec2006}. Since in local time windows, points are not very far away from each other, the algorithm always converges. Thus, given a manifold-valued time series $\gamma(t)$, and a window of length $W$, we compute the mean of the points in the window and this gives rise to the piece-wise aggregate approximation for manifold sequences. When we consider vectors in $\mathbb{R}^n$, this reduces to finding the standard mean of $W$ $n$-dimensional vectors. 
\vspace{-5pt}
\subsection{Symbolic approximation}
\label{subsec:ep_algo}
As discussed above, one of the key-steps in performing symbolic approximation for manifold-valued time-series is to obtain a set of discrete symbols. 
An established theoretical result within the data mining literature is that the efficiency of string searching is optimized when the elements of the codebook are equiprobable \cite{Allauzen2000,devroye92}. The authors of SAX \cite{LinKLC03} emphasize on using equi-probable symbols because they achieve optimal results for fast searching and retrieval using suffix trees, hashing, and Markov models. However, standard clustering approaches do not necessarily result in equiprobable distributions of their centers \cite{zador82,kohonen95,ripley96}. It is also known that when symbols are not equiprobable, there is a possibility of inducing a probabilistic bias in the process \cite{LinL10}. We outline the methods to obtain symbols next.

\subsubsection{Geometry aware K-means for learning symbols}
As a baseline, we chose K-means because it is the most widely used clustering approach and its extension to non Euclidean spaces is well understood. For a set of points $D = (U_1,U_2,\dots,U_n)$ we seek to estimate clusters $\mathbb(C) = (C_1,C_2,\dots,C_K)$ with centers $(\mu_1,\mu_2,\dots,\mu_K)$ such that the sum of geodesic-distance squares, $ \Sigma_{i=1}^{K}\Sigma_{U_j \in C_i}d^2(U_j,\mu_i)$ is minimized. Here $d^2(U_j,\mu_i) = |\mbox{exp}_{\mu_i}^{-1}(U_j)|^2$, where $\mbox{exp}^{-1}$ is the inverse exponential map as described in section \ref{sec:manifoldsPrelim}. We later show that one does not obtain equiprobable symbols using K-means.

\subsubsection{Conscience based competitive learning on manifolds}
To generate symbols or prototypes that divide the feature manifold into equiprobable regions, we extend ideas from Desieno's competitive learning mechanism \cite{Desieno1988} to make it adaptive to the geometry of the space and generate equiprobable symbols. It has been observed that a `conscience' based competitive learning approach does result in symbols that are much more equiprobable than those obtained from clustering approaches. However, the algorithm described in \cite{Desieno1988} is devised only for vector-spaces. Here, we present a generalization of this approach to account for non-Euclidean geometries.

The conscience mechanism starts with a set of initial symbols/prototypes. When an input data-point is presented, a competition is held to determine the symbol closest in distance to the input point. Here, we use the geodesic distance on the manifold for this task. Let us denote the current set of $K$ symbols as $\{S_1, S_2, \ldots, S_K\}$, where each $S_i \in \mathcal{M}$. Let the input data point be denoted as $X \in \mathcal{M}$. The output $y_i$ associated with the $i^{th}$ symbol is described as 

\begin{align}
y_i &= 1, \textrm{ if } d^2(S_i, X) \leq d^2(S_j,X), \forall j \neq i\\
\nonumber y_i &= 0, \textrm{otherwise}
\end{align} 

where, $d()$ is the geodesic distance on the manifold.  Since this version of competition does not keep track of the fraction of times each symbols wins, it is modified by means of a bias term to promote more equitable wins among the symbols. A bias $b_i$ is introduced for each symbol based on the number of times it has won in the past. Let $p_i$ denote the fraction of times symbol $i$ wins the competition. This is updated after each competition as 
\begin{align}
p_i^{new} = p_i^{old} + B(y_i - p_i^{old})
\end{align}  

where $0 < B << 1$. The bias $b_i$ for each symbol is computed as $b_i = C(\frac{1}{K} - p_i)$, where $C$ is a scaling factor chosen to make the bias update significant enough to change the competition (see below). The modified competition is given by

\begin{align}
z_i &= 1, \textrm{ if } d^2(S_i, X) - b_i \leq d^2(S_j,X) - b_j, \forall j \neq i\\
\nonumber z_i &= 0, \textrm{otherwise}.
\end{align}

Finally, the winning symbol is adjusted by moving it partially towards the input data point. The key extension of this algorithm from vector space to non Euclidean spaces lies in this step. In the vector-space version this step is achieved by $S_i^{new} = S_i^{old}+\alpha((X) - S_i^{old}) z_i$.
The partial movement of a symbol towards a data-point can be achieved by means of the exponential and inverse-exponential map as 

\begin{align}
S_i^{new} = \exp_{S_i^{old}} [ \alpha  \exp^{-1}_{S_i^{old}}(X) z_i].
\end{align}
The proposed algorithm for conscience based equi-probable symbol learning is summarized in algorithm \ref{Alg:Conscience}. 

\begin{algorithm}[!htb]
\caption{Equiprobable symbol generation on manifolds.}
\label{Alg:Conscience}
\begin{algorithmic}
\STATE Input: Dataset $\{X_1, \ldots, X_n\} \in \mathcal{M}$. Initial set of symbols $\{S_1, \ldots, S_k\}$.
\STATE Parameters: Biases $b_i = 0$, learning rate $\alpha$, win update factor $B$, conscience factor $C$.

\WHILE{$iter \leq maxiter$}
\FOR{$ j = 1 \to n$}

\STATE $\tilde{i} \leftarrow \min_{i} d^2(X_j,S_i) - b_i$

\STATE $z_{\tilde{i}} = 1$, $z_{i} = 0, i \neq \tilde{i}$

\STATE $S_{i} \leftarrow \exp_{S_i} [ \alpha  \exp^{-1}_{S_i}(X_j) z_i]$

\STATE $p_i \leftarrow p_i + B(z_i - p_i)$

\STATE $b_i \leftarrow C(1/k - p_i) $

\ENDFOR

\ENDWHILE 

\end{algorithmic}
\end{algorithm}

\begin{algorithm}[!htb]
\caption{Symbolic Approximation for Feature Sequences in Euclidean \& Non Euclidean Spaces.}
\label{Alg:SAX}
\begin{algorithmic}

\STATE Input: Feature sequence $\{\beta_1, \ldots, \beta_N\}\in \mathcal{M}$, Learned dictionary $\{D_1, \dots, D_K\}$, Metric $d_\mathcal{M}$ defined on $\mathcal{M}$

\STATE Parameters: Size of aggregating window $W$ ($<<N$),

\STATE Output: Symbolic approximation, {\bf S}.

\STATE $M \leftarrow \ceil{\frac{N}{W}}$.

\STATE $n = 1$

\FOR{$m = 1 \to M$}

	\STATE A$_m \leftarrow {\bf intrinsic~mean}\{\beta_{n}, \beta_{n+1} \dots \beta_{n+W-1}\}$

	\STATE ${\bf S}(m) \leftarrow \underset{1\le j\le K}{\operatorname{argmin}}~~d_\mathcal{M}(A_m,D_j).$
	
	\STATE $n =n+m\times W$

\ENDFOR
\end{algorithmic}
\end{algorithm}

\begin{figure*}[!htb]
 \centering
\subfigure[ K-Means] {\label{kmeans}\includegraphics[clip = true,trim=20mm 15mm 25mm 15mm,width = 2in]{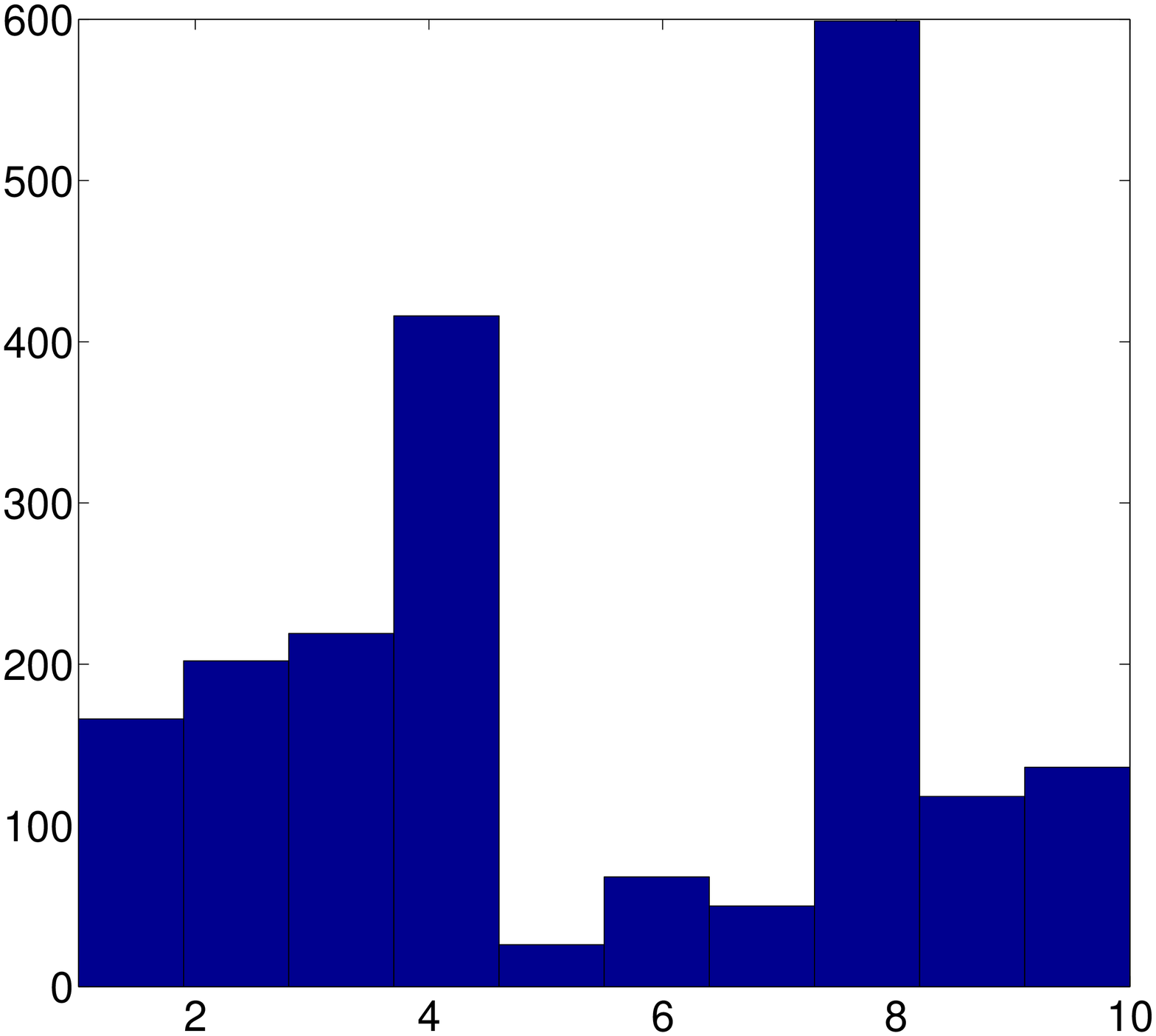}}
\subfigure[ Affinity Propagation]{ \label{aff_prop}\includegraphics[clip = true,trim=20mm 15mm 25mm 15mm,width = 2in]{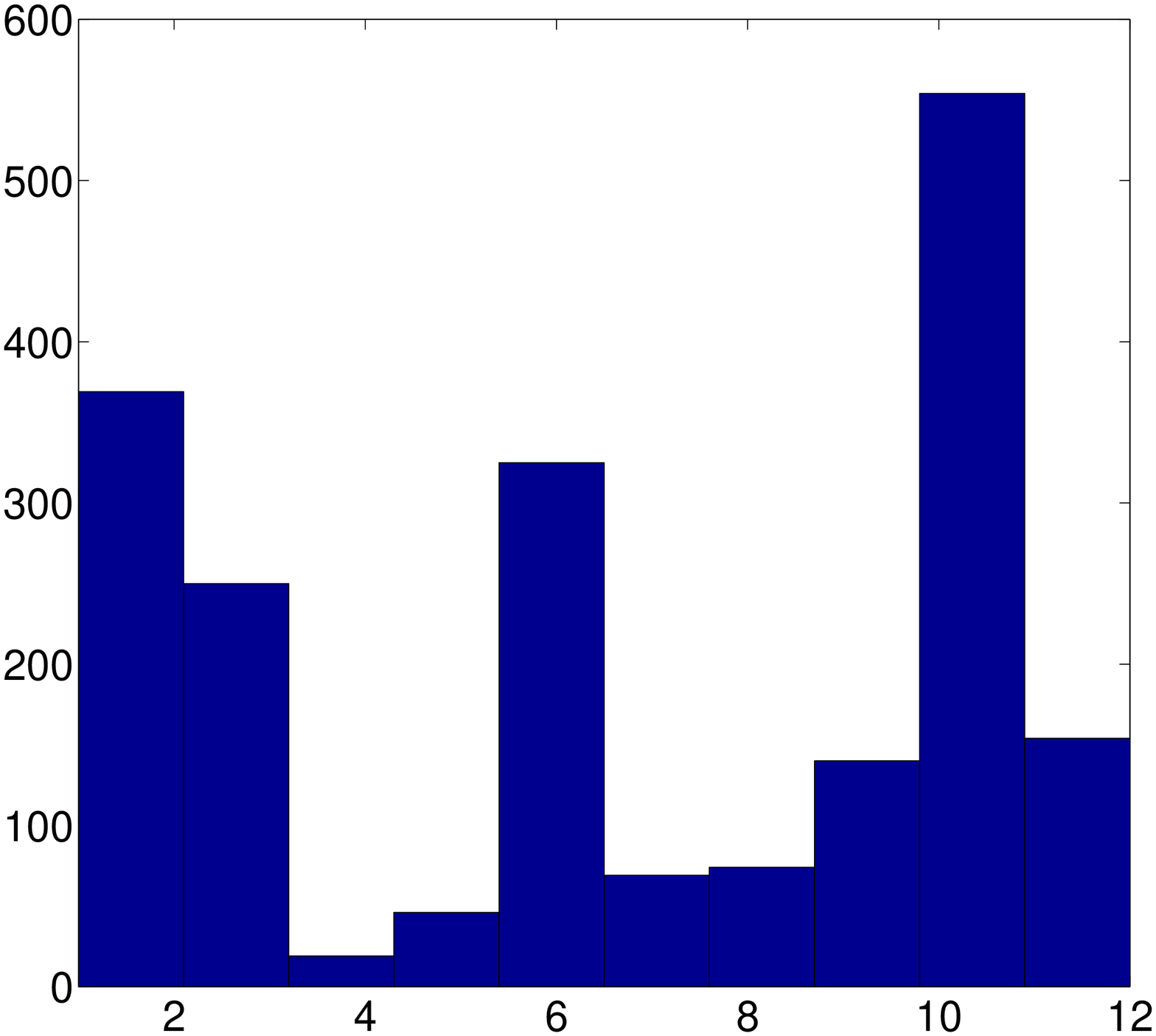}}
\subfigure[Equiprobable Clustering]{ \label{equi_prob}\includegraphics[clip = true,trim=20mm 15mm 25mm 15mm,width = 2in]{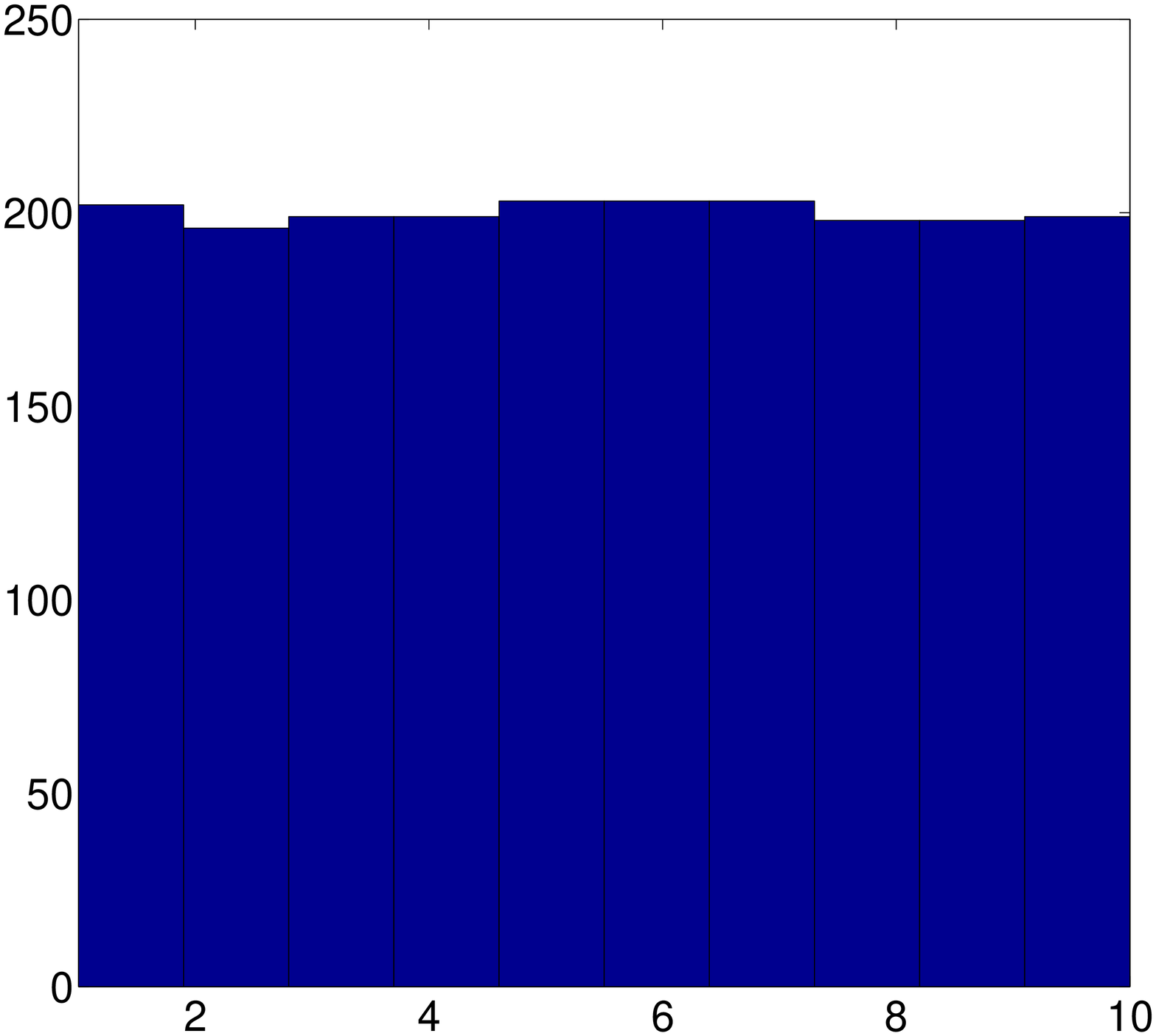}}
    \caption{\small{Probability density functions of the labels generated using (a) K-Means clustering, (b) Affinity Propagation and (c) Equi-Probable Clustering are shown, the feature space in this case was the Grassmann manifold as described in the text. As seen above, equiprobable clustering assigns all clusters with almost equal probability.}}
    \label{hist_clusts}
    \vspace{-5pt}
\end{figure*}
\begin{figure}[!htb]
    \centering
\includegraphics[clip = true,trim=15mm 5mm 20mm 10mm,width = 85mm]{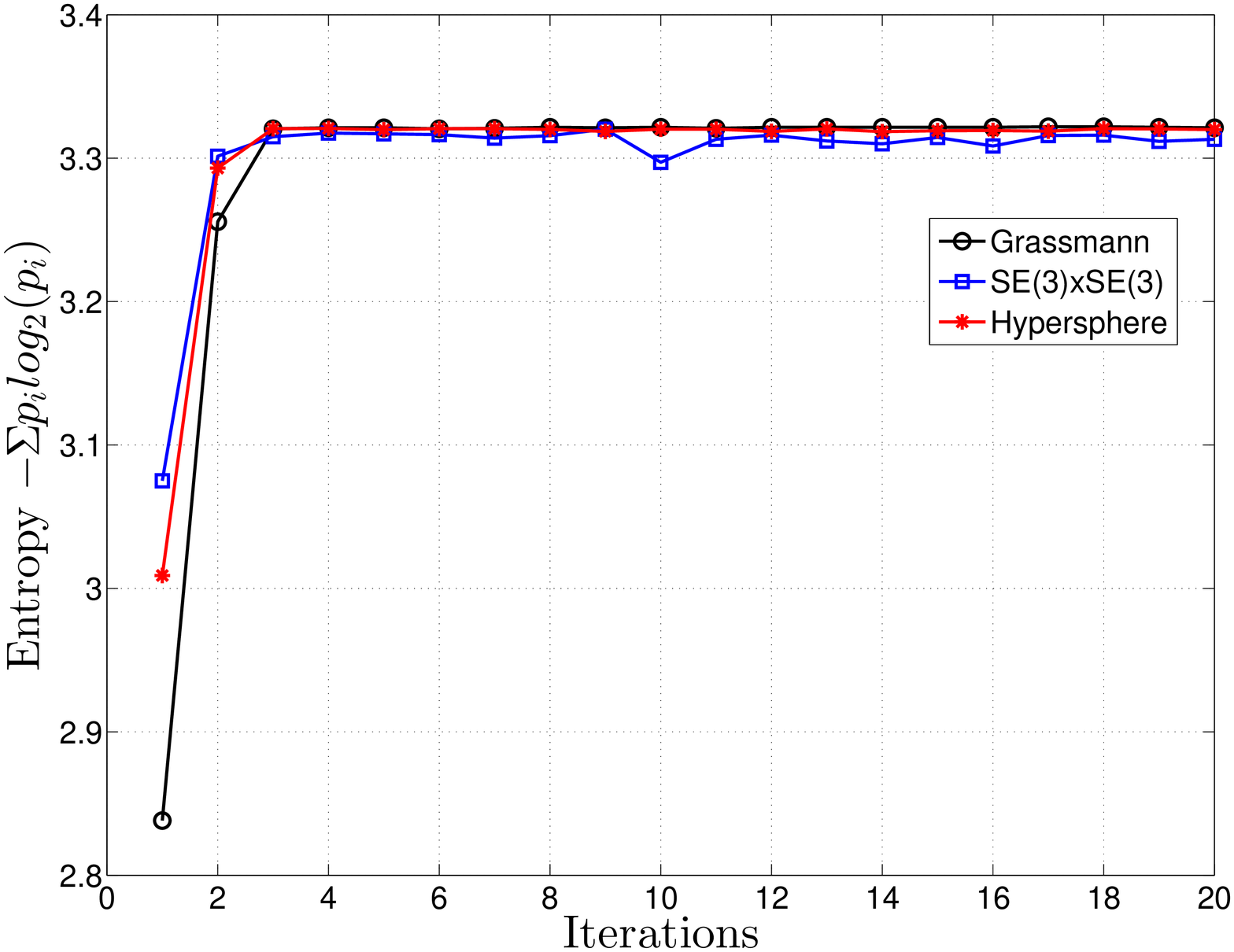}
    \caption{\small{Convergence for the algorithm \ref{Alg:Conscience} on different feature manifolds to obtain $10$ symbols - Grassmannian (UMD), Hypersphere (Weizmann) and $SE(3) \times .. \times SE(3)$(UTKinect). Entropy is plotted as a measure of equiprobability, higher the better.}}
    \label{fig:equi_convergence}
\end{figure}

Next, we illustrate the strength of this approach in obtaining equiprobable symbols on manifolds. For this experiment we chose the UMD human activity dataset \cite{VeeraraghavanC06} and pre-processed it such that we obtain the outer contour of the human. A detailed discussion of the dataset, processing, choice of shape metrics etc. appears in the experiments section. Here, we performed clustering of $2000$ shapes from the dataset into $10$ clusters. We show the histograms of the symbols in fig \ref{hist_clusts}. As seen, both K-means and affinity propagation result in symbols that are far from equiprobable. The proposed approach results in symbols which are much closer to a uniform distribution. The entropy defined as $-\sum_{i = 1}^N p_i \mbox{log}_2(p_i)$, is shown for three different datasets in fig \ref{fig:equi_convergence}. It is seen that the algorithm converges quickly in all cases. Once the symbols are obtained, transforming the feature sequence to its symbolic form is performed using algorithm \ref{Alg:SAX}. 


In practice, while K-means minimizes approximation error it does not have the favorable property of equiprobability, and competitive learning gives us symbols which are equally likely, while compromising on approximation error. In order to find a trade-off between the two, we use a hybrid approach that first uses K-means and then competitive learning from which equiprobable symbols can be obtained in a two stage process. In the first stage we cluster the data using K-means into a small number of clusters, this ensures most data points are adequately represented. Each of these clusters is further split into smaller, equiprobable {\it sub-clusters} in the second stage using conscience learning. The number of clusters in the first stage is an empirical choice, we used values in the range of 5 to 10 for each data set. The number of sub-clusters in the second stage varies according to the probability of their parent cluster. For example, if $p_s$ was the probability of the smallest cluster and we decide to split it into $r$ smaller sub-clusters, then the $i^{th}$ cluster with probability $p_i$ would be split into $\ceil{\frac{p_i}{p_s}\times r}$ clusters. The parameter $r$ indirectly controls the size of the final set of symbols, we used values of $r$ in the range of 1 to 5. We chose these values to obtain a codebook of size($\sim 40-50$). The training phase is expected to be computationally intensive, however this needs to be done only once and can be performed offline and does not affect the speed of comparisons during testing. 
\vspace{-10pt}
\subsection{Limitations and special cases}
Here, we discuss the limitations and some special cases of the proposed formulation. The overall approach assumes that a training set can be easily obtained from which we can extract the symbols for sequence approximation. In the 1D scalar case, this is not an issue, and one assumes that data distribution is a Gaussian, thus the choice of symbols can be obtained in closed-form without any training. If data is not Gaussian, a simple transformation/normalization of the data can be easily performed. In the manifold case, there is no simple generalization of this idea, and we are left with the option of finding symbols that are adapted for the given dataset. For the special case of $\mathcal{M} = \mathbb{R}^n$, the approach boils down to familiar notions of piece-wise aggregation and symbolic approximation with the additional advantage of obtaining data-adaptive symbols, this ensures that the proposed approach is applicable even to the vast class of traditional features used in video analysis.
 For the case of manifolds implicitly specified using samples, we suggest the following approach. One can obtain an embedding of the data into a Euclidean space and apply the special case of the algorithm for $\mathcal{M} = \mathbb{R}^n$. The requirement for the embedding here is to preserve geodesic distances between local pairs of points, since we are only interested in ensuring that data in small windows of time are mapped to points that are close together. Any standard dimensionality reduction approach \cite{Tenenbaum00ISOMAP,RoweisLLE00} can be used for this task. However, recent advances have resulted in algorithms for estimating exponential and inverse exponential maps numerically from sampled data points \cite{Lin2008}. This would make the proposed approach directly applicable for such cases, without significant modifications. Thus the proposed formalism is applicable to manifolds with known geometries as well as to those whose geometry needs to be estimated. 

\section{Speed up in sequence to sequence matching using symbols: applications in activity recognition and discovery} 
\label{sec:speedup}
\begin{figure}[htb!]
    \centering
\includegraphics[clip = true,trim=130mm 0mm 90mm 0mm,width = 85mm]{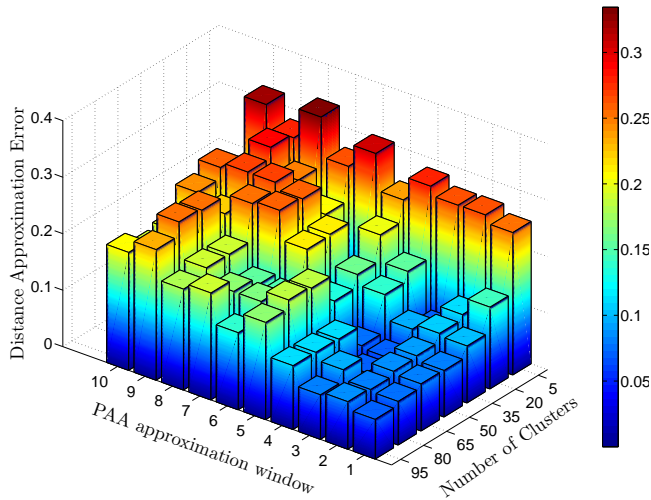}
    \caption{\small{The trade-off between piece-wise aggregation and symbolic approximation is depicted here comparing the error in approximating the distance between two sequences from the Weizmann dataset. A symbol dictionary size of at least $40$ and a approximation window size of up to $3$ has negligible approximation error.}}
    \label{fig:paa_tradeoff}
\end{figure}
The applications considered in this paper are recognition and discovery of human activities. For recognition, a very commonly used approach involves storing labeled sequences for each activity, and performing recognition using a distance-based classifier, a nearest-neighbor classifier being the simplest one. When activity sequences involve manifold-valued time-series, distance computations are quite intensive depending on the choice of metrics. We explore here the utility of the symbolic approximation as an alternative way for approximate yet fast recognition of activities that can replace the expensive geodesic distance computations during testing. As we will show in the experiments, this is especially applicable in real-time deployments and in cases where recognition occurs remotely and there is a need to reduce the communication requirements between the sensor and the analysis engine. Before getting into the details of our experiments and distance metrics used, we define some of the terms used in this paper:
\begin{itemize}
\item[1.]{\it Activity -} In this paper, we will consider an activity to be a high dimensional time series consisting of $N$ data points such that each data point is a feature extracted per frame of the original video. The features can be either Euclidean or belong to abstract spaces such as Riemanian manifolds. We consider cases where all activities may not be of equal lengths by using DTW as a distance metric.
\item[2.]{\it Subsequence -} A subsequence is defined as a contiguous subset of the larger time series, i.e. for a time series $T = (t_1,t_2,\dots,t_n)$ a subsequence of length $n$ is $T_{i,n} = (t_i,t_{i+1},\dots,t_{i+n-1})$.
\item[3.]{\it Motif Discovery -} a pattern that repeats often within a larger time series is known as a motif. We say two patterns within the time series are similar if they are at a distance smaller than some threshold.
\item[4.]{\it Trivial Match -} Within a time series $T$, we say two subsequences $P$ at position $p$ and $Q$ at position $q$ are a trivial match if, $p \in (q-m+1,\dots,q,\dots,q+m-1)$ i.e  $p$ and $q$ are different and within the neighborhood (as specified by $m$) of each other.
\end{itemize}
For an Activity of length $N$, we extract a symbolic representation in windows of size $W$ (where typically $W<<N$). To replace geodesic distance computations for recognition, we will consider subsequences in their symbolic representations to calculate the distance between activities. Let $p_{sub}$ (eg: `bccdea') and $q_{sub}$ (eg: `afffec') be two such subsequences of length $l$, then the distance metric $d_{symbol}$, defined on symbols, is:
\begin{align}
d_{symbol}(p_{sub},q_{sub})=\sum_{i=1}^{l} d_\mathcal{M}\Big ({D}\big(p_{sub}(i)\big),{D}\big(q_{sub}(i)\big)\Big) \label{eqn:symbol_distance}
\end{align}
where $d_\mathcal{M}$ is the metric defined on the manifold, ${ D}$ is the set of symbols or dictionary that is previously learned and ${D}(a)$ is the point on the manifold corresponding to the symbol $a$. Here we assume that the two sequences are of the same length, in other cases we use DTW as a metric or learn a dynamical model for each sequence and use the distance between them as a metric. Since the symbols are known apriori, the distance between them can be computed offline as part of training and stored as a look-up table of pairwise distances between symbols. This allows us to compute distances between sequences in near constant time, which is much faster than computing distances each time using DTW on actual features. 

Before considering applications for the simplified distance measure, one must consider the trade-off between piecewise aggregation, number of symbols versus the error of approximation, this is shown in figure \ref{fig:paa_tradeoff}.

For activity discovery, we consider the problem as one of mining for motifs in time-series. In finding motifs, it is important to consider only non-trivial matches, for every such match we store its location and find the top $k$ motifs. For each of the $k$ motifs, we define a {\it center} for the motif as the sequence which is at minimum distance to all the sequences similar to it. These centers are the $k$ most recurring patterns in the multidimensional time series. We use the brute-force algorithm given in \cite{Patel2002} to extract our motifs.
\section{Experimental Evaluation}
\label{sec:expt}
\begin{figure}[htb!]
    \centering
\includegraphics[clip = true,trim=10mm 30mm 0mm 55mm,width = 3in]{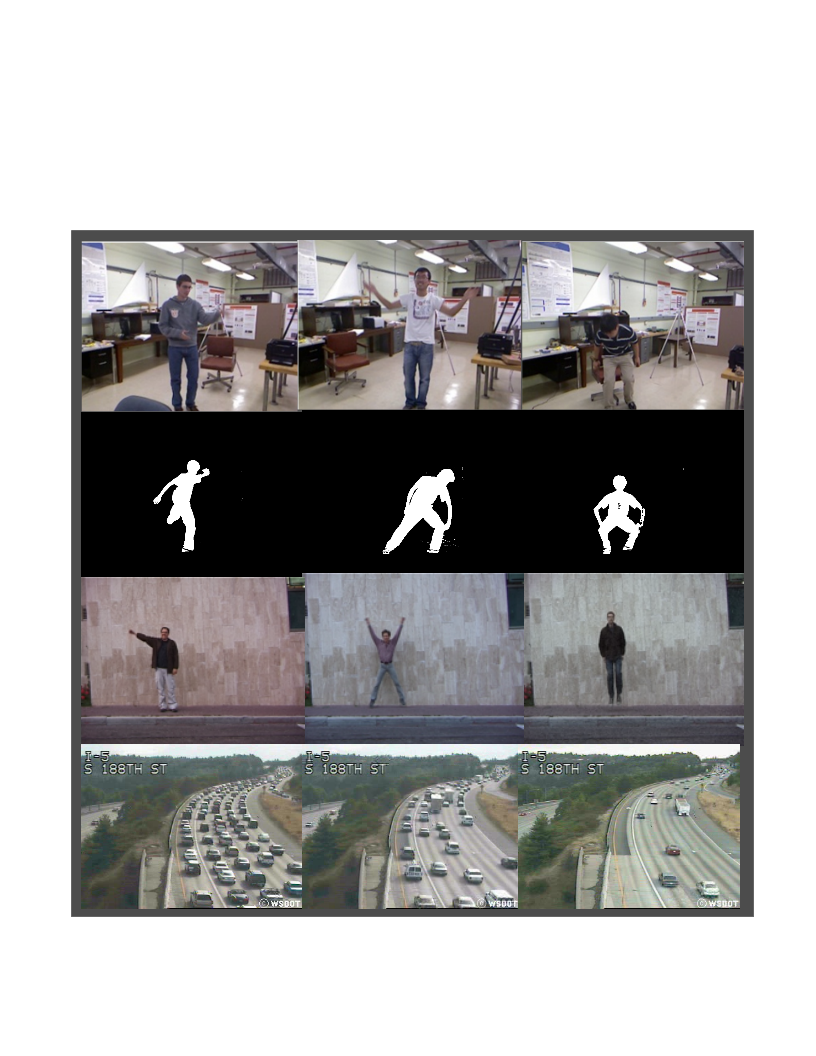}
    \caption{\small{Sample images from the various data sets used in this paper. The UTKinect \cite{xia2012view}, UMD \cite{ashokPAMI05}, the Wiezmann \cite{gorelickpami2007}, and the UCSD traffic \cite{chanTraffic2005} data sets  are shown here from top to bottom in that order.}}
    \label{fig:dataset_sample}
    \vspace{-10pt}
\end{figure}
In this section, we demonstrate the utility of the proposed algorithms for symbolic approximation and its application to activity recognition and discovery. We also study the complexity advantage in using these symbols as compared to original feature sequences. We first describe the datasets and choice of features. 


{\bf UTKinect dataset} \cite{xia2012view} contains 10 activities by 10 subjects, where each activity is repeated twice. There are a total of 199 action sequences. Here we use the feature proposed recently in \cite{VemulapalliCVPR2014}, which models each skeleton as a point on the cross product space of $SE(3) \times \dots \times SE(3)$.

{\bf The UMD database} consists of 10 different activities like bend, jog, push, squat etc.\cite{ashokPAMI05}, each activity was repeated $10$ times, so there were a total of $100$ sequences in the dataset. The background within the UMD Dataset is relatively static which allows us to perform background subtraction. From the extracted foreground, we perform morphological operations and extract the outer contour of the human. We sampled a fixed number of points on the outer contour of the silhouette to yield landmarks, which are represented as points on the Grassmann manifold.

 {\bf The Weizmann Dataset} consists of 93 videos of 10 different actions each performed by 9 different persons \cite{gorelickpami2007}. The classes of actions include running, jumping, walking, side walking etc. Here, the HOOF features \cite{chaudhrycvpr2009} are represented as points on a hyper-spherical manifold.
 
 {\bf The UCSD traffic database} consists of 254 video sequences of daytime highway traffic in Seattle in three patterns i.e. heavy, medium and light traffic \cite{chanTraffic2005}. It was collected from a single stationary traffic camera over two days. 
 
\begin{table}[!htb]
\centering
\begin{tabular}{|p{1.5in}|p{1.3in}|}
	\hline
	Step& Complexity \\
	\hline
	 Exponential map for $\mathcal{M}$ (manifold specific)& O($\nu$) \\
	 \hline
	 Inverse exponential map for $\mathcal{M}$ (manifold specific)&O($\chi$)  \\
	\hline
	 Intrinsic K-means clustering & O((N$\chi$+K$\nu$) $\Gamma$) \\
	\hline
	 Equi-probable clustering& O((NK$\chi$+N$\nu$)$\Gamma$)\\
	\hline
	Approximation of N-length activity to M symbols &  O(M(w$\chi$+$\nu$)$\Gamma$+MK$\chi$)\\
	\hline	
	Symbolic DTW & O($\mbox{M}^2\delta$)\\
	\hline
	Geodesic distance DTW & O($\mbox{M}^2\chi$), $\chi>>>\delta$ \\
	\hline
	\end{tabular}
	\caption{\small{Theoretical complexity analysis for the proposed algorithms. Notations used: N - number of data points, K - number of symbols, with O($\delta$) the time required to read from memory, $\Gamma$ maximum number of iterations, M and w are as defined in algorithm \ref{Alg:SAX} and are usually much lesser than N. It can be seen that a huge complexity gain is achieved in using symbols over original features.}}
	\label{tab:bigO}
	\vspace{-.2in}
\end{table}

\subsection{Speed up and compression achieved using symbols}
\label{sec:speed}
A theoretical complexity analysis of the algorithm is shown in table \ref{tab:bigO}. We  also consider three metrics to study the time-complexity of the proposed framework. Namely 1) Time complexity of matching using symbols vs original feature sequences, 2) Time required to transform a given activity into a symbolic form, and 3) Number of bits required to store/transmit symbols as compared to feature sequences. Ideally, we require that the matching time be several orders of magnitude faster than using the original sequences, the transformation time to be small enough to enable real-time approximation, and very small bit-rate/storage requirement compared to original feature sequences. We show in the following that the proposed framework successfully satisfies all these criteria. We performed the experiments using MATLAB, on a PC with an i7 processor operating at 3.40Ghz with 16GB memory on Windows 7.

\begin{figure*}[!hbt]
  \subfigure[Matching times]{
\centering  
  \includegraphics[clip = true,trim=45mm 10mm 180mm 30mm,scale = 0.25]{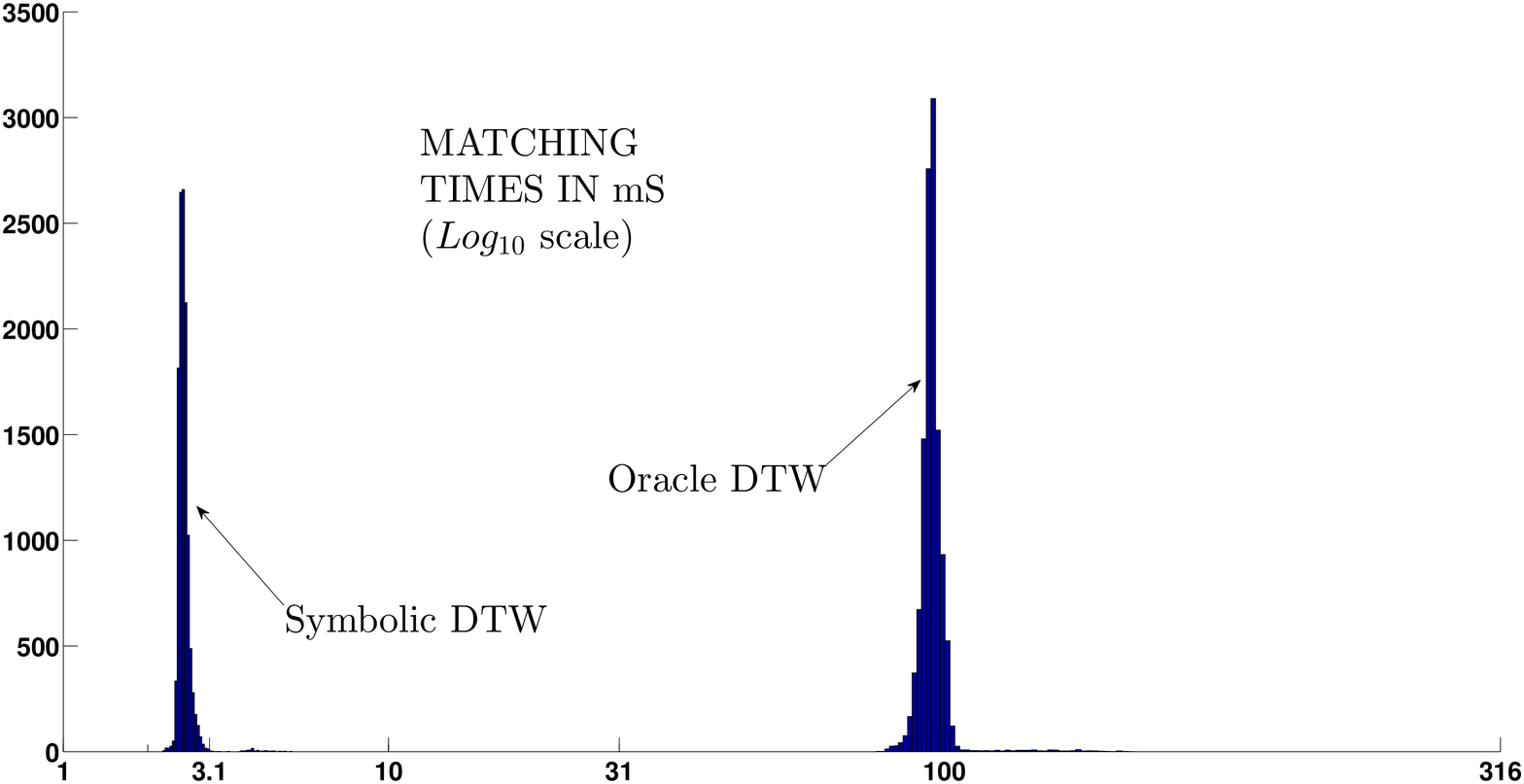}
    \label{fig:match_time}
   }
\subfigure[Converting features to symbols]{
\centering
  \includegraphics[clip = true,trim=45mm 5mm 150mm 30mm,scale = 0.25]{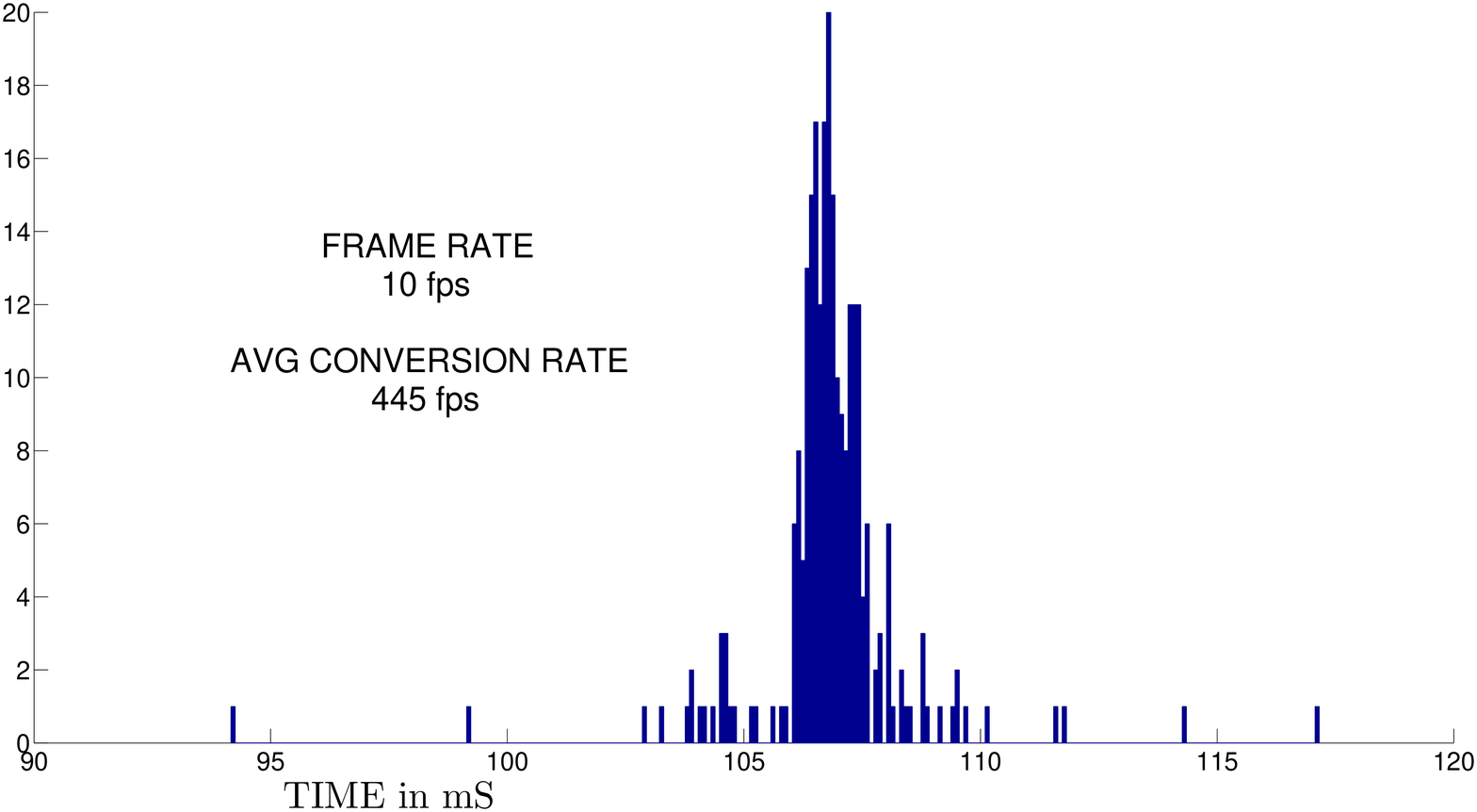}
    \label{fig:symb_translate}
 }
 \subfigure[kNN searching times]{
 \begin{tabular}{|c|c|c|c|}
\hline
Manifold & manifold SAX&Geodesic\\
\hline
Grassmann & $0.0082 \pm 0.0002$s&$1.54 \pm 0.10$s \\
Sphere & $0.24 \pm 0.08$s&$1.22 \pm 0.39$s \\
$SE(3)\times SE(3) ..$ & $0.50 \pm 0.04$s &$57.80 \pm 1.58$s\\\hline
 \end{tabular}
  \label{tab:knnsearch}
 }
 \subfigure[Bit budgets for original vs manifold SAX]{
 \begin{tabular}{|c|c|c|c|} \hline 
{Feature} & { Original Data} & {manifold SAX}&{ Compression}\\ \hline
 
Shape  & 500 Kb & 0.468 Kb &{$\mathbf{99.90\%}$}\\
HOOF feature& 65.625 Kb& 0.410 Kb& {$\mathbf{99.37\%}$}\\
Skeleton feature&  4,617 Kb &111.67 Kb &{$\mathbf{97.58\%}$}\\
\hline
\end{tabular}
\label{tab:bit_budget}
 }
%
     \caption{\small{Comparison of histograms for matching times when using symbolic v/s original feature sequences are shown in fig \ref{fig:match_time} for the UCSD traffic dataset. The times are shown in milliseconds on a log scale. As it can be seen, using symbols speeds up the process by nearly two orders of magnitude. Fig \ref{fig:symb_translate} shows a histogram of times taken to translate entire activities of 50 frames into symbols from the UCSD dataset. Table \ref{tab:knnsearch} shows the improvements in performing a k-NN search on different feature manifolds. Finally table \ref{tab:bit_budget} shows the reduced storage requirements for different features.}}
\end{figure*} 

\subsubsection{kNN search and sequence matching time analysis}
In this experiment we show the gain in speed and compression achieved using symbols compared to using the original high-dimensional features with accompanying metrics. For the gain in speed, we measured the run-time of matching sequences using DTW on symbols vs geodesic DTW. As shown in fig \ref{fig:match_time}, the time taken to match two activity sequences using symbols is just 3.1ms which is two orders of magnitude faster than 100ms that it takes using the actual features. Next, we compare the times taken to perform a k-nearest neighbor (kNN) search on different manifolds in table \ref{tab:knnsearch}. Similar to the sequence matching speed, the search speed is improved by nearly two orders of magnitude.

\subsubsection{Analysis of approximation time}
Fig \ref{fig:symb_translate} shows the distribution of times taken over various activities to transform them into their respective symbolic forms. The average conversion time for an entire activity video is about 107ms. In other words, we can process the video at a speed of 445 frames per second (fps) which allows for easy real time implementation since most videos are recorded at 10-30fps. 
\subsubsection{Bit-rate analysis}
Next, to demonstrate the gain in compression we compared our representation to a baseline using the original feature sequence. Assuming each dimension of the feature is coded as a 32-bit float number, we calculated the bits it would take to represent each feature and its symbolic representation. As shown in {table \ref{tab:bit_budget}}, on nearly all the feature types, the compression ratios are $97\%$ or higher. For a dictionary of size $K$, the number of bits required to represent each symbol is $Log_2(K)$. This provides enough flexibility for the user to choose the size of the codebook and pick features of their choice without significantly affecting the bit-rate.

\subsection{Activity discovery experiment} 
Having learned the symbols, we test their effectiveness in activity discovery. For this experiment, we randomly concatenated $10$ repetitions of $5$ different activities of the UMD dataset to create a sequence that was $50$ activities long. Each activity consists of $80$ frames which were sampled by a sliding window of  size 20 frames with step size of 10 frames. After symbolic approximation, this resulted in $6$ symbols per activity, chosen from an alphabet of 25 symbols. The {\it motifs} or repeating patterns, in five activities - {\it Jogging, Squatting, Bending Knees, Waving and Throwing} were discovered automatically using the proposed method. Each of the discovered motifs was validated manually to obtain a confusion matrix shown in table \ref{tab:confusion_discovery}.  As can be seen, it shows a strong diagonal structure, which indicates that the algorithm works fairly well. Even though all executions of the same activity are not found, we do not find any false matches either.
\begin{table}[htb!]
\begin{center}
\begin{tabular}{|p{0.5in}|p{0.3in}|p{0.3in}|p{0.3in}|p{0.3in}|p{0.3in}|}
\hline
{Activity Type} &  {1} & {2} & {3} & {4} &{5} \\
\hline
{1} & 7 & 0 & 0 & 0 & 0 \\
\hline
{2} & 0 & 7 & 0 & 0 & 0\\
\hline
{3} & 0 & 0 & 8& 0 & 0\\
\hline
{4} & 0 & 0 & 0 & 9 & 0\\
\hline
{5} & 0 & 0 & 0 & 0 & 8\\
\hline
\end{tabular}
\caption{\small{Confusion matrix for the discovered motifs on the UMD database using the manifold SAX representation of the shape feature. Due to the symbolic representation, search can be performed very quickly. Actions discovered are - {\it jogging, squatting, bending, waving and throwing} respectively.}}
\label{tab:confusion_discovery}
\end{center}
     \vspace{-0.3in}

\end{table}
\subsection{Activity recognition using symbols } Symbolic approximation plays a significant role in reducing computational complexity since it allows us to work with symbols instead of working with high dimensional feature sets. In this experiment, we test the utility of the proposed symbolic approximation method for fast and approximate recognition of activities over three datasets. For each data set picking the number of symbols, $K$ is an empirical choice, typically we picked $K = K_{min}$ where, for all $K>K_{min}$ the recognition performance shows no improvement. We also picked a window size of $W = 1$ in our recognition experiments to achieve best performance. A detailed comparison between the window size, number of symbols and performance is seen in figure \ref{fig:paa_tradeoff}, which shows the the error in the geodesic distance vs symbolic distance. To effectively demonstrate the quality of the approximation, we use the classifiers that were reported in the papers that proposed the features. For example, for the shape and the HOOF features, we use the nearest neighbor classifiers, and for the LARP features, we use the SVM. As a baseline, we compare the recognition accuracy of principal geodesic analysis (PGA) \cite{Fletcher2004}, for diffenrent manifolds.
\begin{table}[!htb!]
\centering
{\small
\begin{tabular}{|p{3.2cm}|l|p{1.5cm}|}
 \hline
Activity & Accuracy ($\%$) & Relative bit budget \\
\hline
Shape + manifold SAX & 98 & 1 \\ 
\hline
Shape + PGA \cite{Fletcher2004}& 90 & 6.012\\ 
\hline
Shape \cite{ashokPAMI05}&100&1202.6\\
\hline
\end{tabular}}
\caption{\small{Recognition experiment for the UMD database with a shape silhouette feature. Here we see the performance achieved with symbolic approximation compared to an oracle geodesic distance based nearest neighbor classifier.}}
\label{tab:recognition_accuracy}
\vspace{-0.1in}
\end{table}

For the UMD dataset, we learned a dictionary of $60$ symbols using algorithm \ref{Alg:Conscience}. Then, we performed a recognition experiment using a leave one-execution-out test in which we trained on $9$ executions and tested on the remaining execution, the results are shown in Table \ref{tab:recognition_accuracy}. It can be seen that the recognition performance using symbols is very close to that obtained by using an oracle geodesic distance DTW based algorithm. We achieve this performance with matching times that are significantly faster, as will be described in section \ref{sec:speed}.  

For the UTKinect dataset, we learn a common alphabet of size $20-25$ symbols for all the relative joints from actions corresponding to the training subjects. The approximated LARP features are then mapped to their corresponding Lie algebra following the protocol of \cite{VemulapalliCVPR2014}. Finally these features are classified using a one-vs-all SVM classifier similar to \cite{VemulapalliCVPR2014}. Here, our results are reported without any post-processing using FTP as done in \cite{VemulapalliCVPR2014}, which improves performance further. Results show that even with a small dictionary size, there is negligible loss in recognition accuracy, while drastically reducing the search speed \ref{tab:knnsearch} by a factor of nearly 50. Even though we approximate the actions with a small codebook, we obtain a better recognition performance than the original features. This is explained by the fact that the Lie algebra, $\mathfrak{se}(3)\in \real^6$, which is much lower than the other features considered here and therefore can be appropximated much better with fewer symbols. The approximated LARP features also provide robustness to noise, which is common in features extracted using Kinect. 


\begin{table}[!htb]
\centering
	\begin{tabular}{|p{1.3in}|l|p{0.5in}|}
	\hline
	Feature & Accuracy($\%$) & Relative bit budget\\
	\hline
	LARP+ manifold SAX & {\bf 94.77}& 1\\
	\hline
	LARP+PGA \cite{Fletcher2004}& 92.46&20.428 \\
	\hline
	LARP \cite{VemulapalliCVPR2014} &92.97& 40.856 \\
	\hline
	HOG3D \cite{xia2012view}& 90.00&NA \\
	\hline
	\end{tabular}
	\caption{\small{Results on the UTKinect dataset.}}
	\label{tab:utkinect}
	\end{table}
For the Weizmann dataset, we demonstrate the flexibility of the approximation strategy by learning linear dynamical models over the approximated sequences, which also serves as a fair comparison to the state of the art techniques. We performed the recognition experiment on all the $9$ subjects performing $10$ activities each with a total of $90$ activities. The dictionary learned had 55 symbols which were used to map the activities to the approximated sequences. Next, we fit a linear dynamical model to the approximately  reconstructed actions and perform recognition with a nearest neighbor classifier using the Martin metric on LDS parameters \cite{Soatto2001}. The results for the leave-one-execution-out recognition test are shown in Table \ref{tab:weizmann1} and it can be seen there is almost no loss in performance in comparison to state of the art techniques. Better results have been reported on this dataset by Gorelick {\it et al.} \cite{gorelickpami2007} etc., but there are no common grounds between their technique or feature and ours for it to be a fair comparison.

\begin{table}[!htb]
\centering
	\begin{tabular}{|p{1.4in}|l|p{0.5in}|}
	\hline
	Feature & Accuracy($\%$) & Relative bit budget\\
	\hline
	 LDS+ manifold SAX & {\bf 92.22}& 1\\
	\hline
	HOOF+DTW+manifold SAX & 88.87&1 \\
	\hline
	HOOF+DTW+PGA \cite{Fletcher2004} & 74.44& 10.67 \\
	\hline
	HOOF+DTW \cite{chaudhrycvpr2009} & 90.00& 160 \\
	\hline
	$\chi^2$-Kernel \cite{chaudhrycvpr2009}  & 95.66& 160\\
\hline
	Chaotic~measures \cite{AliBS07} & 92.60& NA\\
	\hline
	\end{tabular}
	\caption{\small{Recognition Performance for the Weizmann dataset.}}
	\label{tab:weizmann1}
	\end{table}
	
\begin{table}[!htb]
\centering
\begin{tabular}{|p{0.5in}|p{0.5in}|p{0.5in}| p{0.5in}|}
	\hline
	 & Manifold SAX ($\%$)& CS LDS($\%$) &Oracle LDS($\%$) \\
	\hline
	 {Expt 1} &  { 84.13} &  85.71 & 77.77 \\
	\hline
	 {Expt 2}& { 82.81}& 73.43 & 82.81\\
	\hline
	{Expt 3}&{79.69} & 78.10 & 91.18  \\
	\hline
	 {Expt 4}& { 79.37} & 76.10 & 80.95  \\
	\hline
	{Average} & {\bf 81.50} & 78.33 & 83.25  \\
	\hline
	\end{tabular}
	\caption{\small{Recognition performance for UCSD traffic data set. The results for Oracle LDS and CS LDS are from \cite{AswinECCV10}.}}
	\label{tab:traffic1}
\end{table}

For the Traffic Database, we stacked every other pixel in the rows and columns of each frame to form our feature vector. We learned 45 symbols from the training set using these features. We performed the recognition experiment on $4$ different test sets which contained ~25\% of the total videos. We used a 1-NN classifier with a DTW metric on the symbols. The results are shown in Table \ref{tab:traffic1}. We compare our results to \cite{AswinECCV10}, which also performed recognition using lower dimensional feature representation using compressive sensing. As it can be seen, recognition performance is clearly better when the feature is in its symbolic form as compared to when it was compressively sensed, given that both are significantly reduced versions of the original feature. We also perform nearly as well as the performance achieved using the original feature itself.
\vspace{-0.2in}
\section{Discussion and Future Work}
\label{sec:discuss}
In this paper we presented a formalization of high dimensional time-series approximation for efficient and low-complexity activity discovery and activity recognition. We presented geometry and data adaptive strategies for symbolic approximation, which enables these techniques for new classes of non-Euclidean visual representations, for instance in activity analysis. The results show that it is possible to significantly reduce Riemannian computations during run-time by an intrinsic indexing and approximation algorithm which allows for easy and efficient real time implementation. This opens several avenues for future work like an integrated approach of temporal segmentation of human activities and symbolic approximation. A theoretical and empirical analysis of the advantages of the proposed formalism on resource-constrained systems such as robotic platforms would be another avenue of research. 

Finally, the framework in this paper is general enough to deal with more abstract forms of information such as graphs \cite{jordan1998} or bag-of-words \cite{GaurZSC11}. In fact, any system that is sequential can be used within this framework, the key is to have a good understanding of metrics on these abstract models. Existing works have defined kernels for data on manifolds \cite{Lafferty2005}, for graphs \cite{Vishwanathan2008} and a good starting point would be to use these to develop a kernel version of this framework that would allow us to learn symbols. 

\bibliographystyle{apalike}
\bibliography{biblio2,biblio1}
\end{document}